%% file: main.tex
\RequirePackage{graphicx}
\documentclass[10pt, conference, ]{IEEEtran}
\IEEEoverridecommandlockouts
\input{preamble}

\input{acronyms}

\begin{document}

\title{Adapting the Behavior of Reinforcement Learning Agents to Changing Action Spaces and Reward Functions}

\author{
\IEEEauthorblockN{Raul de la Rosa}
\IEEEauthorblockA{
\textit{Universidad de los Andes}\\
Bogot\'a, Colombia \\
c.delarosap@uniandes.edu.co}
\and
\IEEEauthorblockN{Ivana Dusparic}
\IEEEauthorblockA{
\textit{Trinity College Dublin}\\
Dublin, Ireland \\
ivana.dusparic@tcd.ie}
\and
\IEEEauthorblockN{Nicolas Cardozo}
\IEEEauthorblockA{
\textit{Universidad de los Andes}\\
Bogot\'a, Colombia \\
n.cardozo@uniandes.edu.co}
}

\maketitle

\begin{abstract}
Reinforcement Learning (RL) agents often struggle in real-world applications where environmental conditions are non-stationary, particularly when reward functions shift or the available action space expands. This paper introduces \adaptiverl, a self-adaptive Q-learning framework that enables on-the-fly  adaptation without full retraining. By integrating concept drift detection with dynamic adjustments to learning and exploration hyperparameters, \adaptiverl adapts agents to changes in both the reward function and on-the-fly expansions of the agent's action space, while preserving prior policy knowledge to prevent catastrophic forgetting. We validate our approach using a Gridworld benchmark and a traffic signal control simulation. The results demonstrate that \adaptiverl achieves superior convergence speed and continuous adaptation compared to a standard Q-learning baseline, improving learning efficiency by up to 1.7x.
\end{abstract}


\input{introduction}

\input{background}

\input{implementation}

\input{validation}

\input{related}

\input{conclusion}


\printbibliography

\end{document}

%% file: preamble.tex
\usepackage{ifdraft}

\usepackage[T1]{fontenc}
\usepackage[utf8]{inputenc}
\usepackage{amsmath}
\usepackage{mathrsfs}
\usepackage{hyperref}
\usepackage{wasysym}
\usepackage{yfonts}
\usepackage[plain]{fancyref}
\usepackage{subcaption}

\usepackage[final]{listings}
\usepackage{booktabs}
\usepackage{multirow}
\usepackage{xcolor,colortbl}
\usepackage{xspace}
\usepackage{suffix}
\usepackage{etoolbox}
\usepackage[inline]{enumitem}
\usepackage{acronym}
\usepackage{url}
\usepackage{textgreek} 
\usepackage{algorithm}
\usepackage{algpseudocode}

\usepackage[backend=bibtex,
   style=ieee,
   citestyle=numeric-comp,
   natbib=true,
   maxnames=7,
   minnames=1,
   maxcitenames=1, 
   mincitenames=1,
   giveninits=true,
   hyperref=true,
   sorting=none,
   defernumbers]{biblatex}

\def\fref{\Fref} 
\renewcommand{\lstlistingname}{Snippet}
\newcommand*{\fancyreflstlabelprefix}{lst} 
\newcommand*{\Freflstname}{\lstlistingname}
\newcommand*{\freflstname}{\lstlistingname}
\Frefformat{vario}{\fancyreflstlabelprefix}%
  {\Freflstname\fancyrefdefaultspacing#1#3}
\frefformat{vario}{\fancyreflstlabelprefix}%
  {\freflstname\fancyrefdefaultspacing#1#3}
\Frefformat{plain}{\fancyreflstlabelprefix}%
  {\Freflstname\fancyrefdefaultspacing#1}
\frefformat{plain}{\fancyreflstlabelprefix}%
  {\freflstname\fancyrefdefaultspacing#1}

\newcommand*{\fancyrefthmlabelprefix}{thm} 
\newcommand*{\Frefthmname}{Theorem}%
\newcommand*{\frefthmname}{%
 \MakeLowercase{\Frefthmname}}%
\Frefformat{vario}{\fancyrefthmlabelprefix}%
  {\Frefthmname\fancyrefdefaultspacing#1#3}
\frefformat{vario}{\fancyrefthmlabelprefix}%
  {\frefthmname\fancyrefdefaultspacing#1#3}
\Frefformat{plain}{\fancyrefthmlabelprefix}%
  {\Frefthmname\fancyrefdefaultspacing#1}
\frefformat{plain}{\fancyrefthmlabelprefix}%
  {\frefthmname\fancyrefdefaultspacing#1}

\newcommand*{\fancyreflemlabelprefix}{lem} 
\newcommand*{\Freflemname}{Lemma}%
\newcommand*{\freflemname}{%
 \MakeLowercase{\Freflemname}}%
\Frefformat{vario}{\fancyreflemlabelprefix}%
  {\Freflemname\fancyrefdefaultspacing#1#3}
\frefformat{vario}{\fancyreflemlabelprefix}%
  {\freflemname\fancyrefdefaultspacing#1#3}
\Frefformat{plain}{\fancyreflemlabelprefix}%
  {\Freflemname\fancyrefdefaultspacing#1}
\frefformat{plain}{\fancyreflemlabelprefix}%
  {\freflemname\fancyrefdefaultspacing#1}

\newcommand*{\fancyrefdeflabelprefix}{def} 
\newcommand*{\Frefdefname}{Definition}%
\newcommand*{\frefdefname}{%
 \MakeLowercase{\Frefdefname}}%
\Frefformat{vario}{\fancyrefdeflabelprefix}%
  {\Frefdefname\fancyrefdefaultspacing#1#3}
\frefformat{vario}{\fancyrefdeflabelprefix}%
  {\frefdefname\fancyrefdefaultspacing#1#3}
\Frefformat{plain}{\fancyrefdeflabelprefix}%
  {\Frefdefname\fancyrefdefaultspacing#1}
\frefformat{plain}{\fancyrefdeflabelprefix}%
  {\frefdefname\fancyrefdefaultspacing#1}

\newcommand*{\fancyreflnlabelprefix}{ln} 
\newcommand*{\Freflnname}{Line}%
\newcommand*{\freflnname}{%
 \MakeLowercase{\Freflnname}}%
\Frefformat{vario}{\fancyreflnlabelprefix}%
  {\Freflnname\fancyrefdefaultspacing#1#3}
\frefformat{vario}{\fancyreflnlabelprefix}%
  {\freflnname\fancyrefdefaultspacing#1#3}
\Frefformat{plain}{\fancyreflnlabelprefix}%
  {\Freflnname\fancyrefdefaultspacing#1}
\frefformat{plain}{\fancyreflnlabelprefix}%
  {\freflnname\fancyrefdefaultspacing#1}

\lstdefinestyle{floating}
 {frame=lines,
  float=hptb,
  captionpos=b,
  abovecaptionskip=-0pt}

\lstdefinestyle{py}
 {language=Python,
  showstringspaces=false,
  keywordstyle=\ttfamily\bfseries,
  tabsize=2,
  style=floating,
  belowskip=-0\baselineskip,
  aboveskip=-0\baselineskip,
  morekeywords={}
}

\lstnewenvironment{python}[1][]
 {\lstset{style=py,#1}}{}  

\usepackage{tikz}
\usepackage[export]{adjustbox}
\usetikzlibrary{positioning,shadows,trees,mindmap}
\usetikzlibrary{arrows, shapes, backgrounds}
\usetikzlibrary{decorations.pathreplacing}
\usetikzlibrary{shapes.misc}
\usepackage[edges]{forest}
\usetikzlibrary{arrows.meta}
\colorlet{linecol}{black!75}
\usepackage{xkcdcolors} 
\usetikzlibrary{tikzmark}
\usetikzlibrary{calc}
\newcommand{\highlight}[2]{\colorbox{#1!17}{#2}}

\newcommand{\eg}{\emph{e.g.,}\xspace}
\newcommand{\ie}{\emph{i.e.,}\xspace}

\newcommand{\adaptiverl}{\textsc{morphin}\xspace}

\definecolor{author}{rgb}{.5, .5, .5}
\definecolor{comment}{rgb}{.1, .0, .9}
\definecolor{note}{rgb}{.9, .4, .0}
\definecolor{idea}{rgb}{.1, .7, .0}
\definecolor{missing}{rgb}{.9, .1, .0}
\definecolor{OliveGreen}{rgb}{0,0.6,0.3}
\definecolor{Bittersweet}{rgb}{0.996,0.435,0.369}
\definecolor{RoyalBlue}{rgb}{0.25,0.41,0.88}
\definecolor{NavyBlue}{rgb}{0.0,0.4,0.8}
\definecolor{Mulberry}{rgb}{0.77,0.29,0.54}

\makeatother

%% file: acronyms.tex
\acrodef{AI}{Artificial Intelligence}
\acrodef{API}{Application Programming Interface}
\acrodef{AOP}{Aspect-oriented Programming}
\acrodef{CAS}{Collective Adaptive Systems}
\acrodef{CRL}{Continual Reinforcement Learning}
\acrodef{CONRL}[ConRL]{Constructivist Reinforcement Learning}
\acrodef{DSL}{Domain-Specific Language}
\acrodef{DNN}{Deep Neural Networks}
\acrodef{DQN}{Deep Q-Networks}
\acrodef{IOT}[IoT]{Internet of Things}
\acrodef{LOC}{Lines of Code}
\acrodef{ML}{Machine Learning}
\acrodef{MDP}{Markov Decision Process}
\acrodef{MNC}{Multiply-Nested Container}
\acrodef{OOP}{Object-Oriented Programming}
\acrodef{PCA}{Principal Component Analysis}
\acrodef{PDG}{Program Dependence Graph}
\acrodef{RL}{Reinforcement Learning}

\newcommand{\acResetNonTrivial}
  {\acresetall
   \acused{CPU}
   \acused{API}
   \acused{LAN}
   \acused{SMT}
   \acused{GUI}}

\acResetNonTrivial

%% file: introduction.tex

\section{Introduction}
\label{sec:introduction}

Traditional \acf{RL} algorithms assume stationary \acp{MDP}, where transition probabilities and reward functions remain constant~\cite{sutton18}. However, real-world environments often exhibit non-stationary characteristics like evolving reward dynamics and changing action spaces~\cite{khetarpal2022continualreinforcementlearningreview}, limiting the effectiveness of standard approaches.

This paper introduces \adaptiverl, a self-adaptive Q-learning algorithm that addresses two key challenges in non-stationary environments: (1) adapting to changing reward functions (goals), and (2) incorporating new actions into the agent's behavior. Our approach integrates concept drift detection using the Page-Hinkley test (PH-test)\cite{changingpointdetection} with dynamic adjustment of learning ($\alpha$) and exploration ($\varepsilon$) rates, enabling agents to explore for enough time in order to preserve prior knowledge \cite{norman2024firstexploreexploitmetalearningsolve} while adapting to environmental changes.

\adaptiverl operates through two main mechanisms: \emph{environment monitoring} for drift detection, and \emph{adaptive learning} that dynamically adjusts parameters based on temporal difference errors. When concept drift is detected, the agent increases exploration until the agents stabilize over the new configuration while maintaining its Q-table structure, allowing rapid adaptation without catastrophic forgetting. For new actions, the agent extends its Q-table dimensions and applies targeted exploration to integrate new capabilities.

We validate our approach on Gridworld benchmarks with shifting goals and expanding action spaces, and demonstrate practical applicability in traffic signal control scenarios. Results show superior convergence speed and adaptation efficiency compared to standard Q-learning baselines, positioning \adaptiverl as a suitable approach for self-adaptive systems operating in dynamic environments.

%% file: background.tex

\section{Background}
\label{sec:background}

\subsection{\acl{RL} in Non-Stationary Environments}

\ac{RL} agents learn optimal policies by interacting with environments formulated as \acp{MDP} $\mathcal{M} = \langle \mathcal{S}, \mathcal{A}, P, R, \gamma \rangle$, where $\mathcal{S}$ and $\mathcal{A}$ are the state and action spaces, $P$ represents transition probabilities, $R$ is the reward function, and $\gamma$ is the discount factor~\cite{sutton18}.

Real-world environments often violate stationarity assumptions, leading to \emph{non-stationary} MDPs represented as temporal sequences:
\[
\{\mathcal{M}_t\}_{t=1}^\infty \quad\text{with}\quad \mathcal{M}_t = \bigl\langle \mathcal{S}, \mathcal{A}_t, P_t, R_t, \gamma \bigr\rangle
\]
where changes in $R_t$ and $\mathcal{A}_t$ constitute \emph{concept drifts}, which refer to shifts in the environment's underlying dynamics. This implies that the agent's previously learned policy and Q-values may become suboptimal for maximizing future rewards. Q-learning updates action values using the Bellman equation:
\vspace{1em}
\input{equations/bellman}

\vspace{1em}

While effective in stationary settings, standard Q-learning struggles when $R_t$ or $\mathcal{A}_t$ change over time. Our work addresses this limitation through adaptive mechanisms that detect and respond to environmental changes in the form of modifications to the reward distribution function and action space expansions. We focus on rewards and action spaces, as changes in the state space has been addressed in the literature~\cite{ConstructivistRL}.

\subsection{Motivating Example}
\label{sec:motivation}
We motivate our work using a $9\times 9$ Gridworld benchmark, where agents navigate from the center of the board to a goal state (\ie a state with reward $+100$) while avoiding negative rewards (states with reward $-1$). The environment presents two adaptation challenges:
\begin{itemize}
    \item Goal Relocation: The target switches between corners, from the top-left corner to the bottom 
    right-corner. This requiring agents to adapt to new reward distributions without losing previously 
    learned knowledge, as the goal changes multiple times. In each change, the agent should take less 
    time in learning the behavior to reach the goal, than the required time to learn from scratch.
    \item Action Space Expansion: New actions (\eg jumping over obstacles) are introduced, requiring 
    agents to integrate new capabilities into existing policies to further optimize the policy with the new 
    actions.
\end{itemize}

%% file: equations/bellman.tex
\begin{equation*} \label{eq:QL}
     {\tikzmarknode{qt}{\highlight{purple}{$Q(s_t, a_t)$}}} +
    {\tikzmarknode{alpha}{\highlight{NavyBlue}{$\alpha$}}}
    [ 
    {\tikzmarknode{r}{\highlight{Bittersweet}{$r_{t+1}$}}}  +  
    {\tikzmarknode{gamma}{\highlight{RoyalBlue}{$\gamma$}}}
    {\tikzmarknode{max}{\highlight{OliveGreen}{$\max\limits_a Q(s_{t+1},a)$}}} - 
    {\tikzmarknode{qt2}{\highlight{purple}{$Q(s_t, a_t)$}}} 
    ]
\end{equation*}

\begin{tikzpicture}[overlay,remember picture,>=stealth,nodes={align=left,inner ysep=1pt},<-]
    \path (qt.north) ++ (3.9,1.7em) node[anchor=south east,color=Mulberry!85] (ntext){\textsf{\footnotesize Q-value}};
    \draw [color=Mulberry](qt.north) |- ([xshift=0.8ex,color=Mulberry]ntext.south west);
    \path (qt2.north) ++ (-2.2,1.8em) node[anchor=south east,color=Mulberry!85] (qt2text){};
    \draw [color=Mulberry](qt2.north) |- ([xshift=-4.9ex,color=Mulberry]qt2text.south west);
    \path (alpha.north) ++ (-0.2,-2.8em) node[anchor=south east,color=NavyBlue] (atext){\textsf{\footnotesize learning rate}};
    \draw [color=NavyBlue](alpha.south) |- ([xshift=-9.3ex,color=NavyBlue]atext.south east);
    \path (r.north) ++ (-0.1,1.5em) node[anchor=north east,color=Bittersweet!85] (lijtext){\textsf{\footnotesize reward}};
    \draw [color=Bittersweet](r.north) |- ([xshift=-4.3ex,color=Bittersweet]lijtext.south east);
    \path (gamma.north) ++ (0.5,1.5em) node[anchor=north west,color=RoyalBlue!85] (gtext){\textsf{\footnotesize discount factor}};
    \draw [color=RoyalBlue](gamma.north) |- ([xshift=-2.9ex,color=RoyalBlue]gtext.south east);
    \path (max.north) ++ (-1.2,-3.6em) node[anchor=south west,color=xkcdHunterGreen!85] (lmaxtext){\textsf{\scriptsize Maximum Q-Value in the next state}};
    \draw [color=xkcdHunterGreen](max.south) |- ([xshift=-5ex,color=xkcdHunterGreen]lmaxtext.north);
\end{tikzpicture}

%% file: implementation.tex

\section{\ac{RL} Agents with Adaptive Behavior}
\label{sec:implementation}

This section introduces \adaptiverl, a self-adaptive Q-learning framework that enables agents to dynamically adapt to non-stationary environments, specifically to changes in reward functions (goals) and on-the-fly expansions of the action space. \adaptiverl integrates two core components, detailed in Algorithm \ref{alg:morphin}: proactive environment monitoring using a concept drift detector, and an adaptive learning process that modulates exploration and learning rates to incorporate new knowledge while preserving prior experience. The implementation is publicly available.\footnote{Available at: \url{https://anonymous.4open.science/r/morphin_rl}} This approach aligns with the principles of \ac{CRL}~\cite{abel2023definitioncontinualreinforcementlearning} and Self-Adaptive Systems (SAS)~\cite{WONG2022106934} by enabling agents to autonomously modify their learning strategy at runtime, ensuring resilience in non-stationary contexts.

\begin{algorithm}[hbt!]
\caption{\adaptiverl: Adaptive Q-Learning with Concept Drift Detection}
\label{alg:morphin}
\begin{algorithmic}[1]
\State \textbf{Initialize} parameters:
\State \quad Base learning rate $\alpha$, max learning rate $\alpha_{\max}$, discount factor $\gamma$
\State \quad TD-error sensitivity $k$, exploration decay parameters $\varepsilon_{\min}, \text{decay\_rate}$
\State \quad Page-Hinkley parameters: sensitivity $\delta$, threshold $H$
\State \textbf{Initialize} Q-table $Q(s, a) \leftarrow 0$ for all $s \in S, a \in A$
\State \textbf{Initialize} drift detector $PH\_Test(\delta, H)$
\State \textbf{Initialize} exploration decay counter $e \leftarrow 0$

\For{episode = 1 to N}
    \State Reset state $s_t \leftarrow s_{\text{initial}}$
    \State Reset cumulative episode reward $R_{ep} \leftarrow 0$
    
    \While{$s_t$ is not terminal}
        \State $\varepsilon_t \leftarrow \varepsilon_{\min} + (1 - \varepsilon_{\min}) \cdot \exp(-\text{decay\_rate} \cdot e)$
        \State $a_t \leftarrow \text{choose\_action}(s_t, \varepsilon_t)$ \Comment{$\varepsilon$-greedy selection}
        \State Execute $a_t$, observe $r_{t+1}$ and $s_{t+1}$
        \State $R_{ep} \leftarrow R_{ep} + r_{t+1}$
        \State\Comment{Adaptively update Q-value with eq. \eqref{eq:td_error} \& \eqref{eq:dynamic_learning_rate}}
        \State $Q(s_t, a_t) \leftarrow Q(s_t, a_t) + \alpha^* \cdot TD_{error}$
        \State $s_t \leftarrow s_{t+1}$
    \EndWhile
    
    \State\Comment{Detect Concept Drift at the end of the episode}
    \If{$PH\_Test.update(R_{ep})$ is True}
        \State \Comment{Drift detected: reset exploration schedule}
        \State $e \leftarrow 0$
        \State $PH\_Test.reset()$
    \Else
        \State \Comment{Stable environment: continue exploration decay}
        \State $e \leftarrow e + 1$
    \EndIf
\EndFor
\end{algorithmic}
\end{algorithm}

\subsection{Environment Monitoring and Adaptation}
\label{sec:morphin-adaptation}

\begin{figure*}[hbtp]
    \centering
    \begin{subfigure}[b]{0.49\textwidth}
        \centering
        \includegraphics[width=\linewidth]{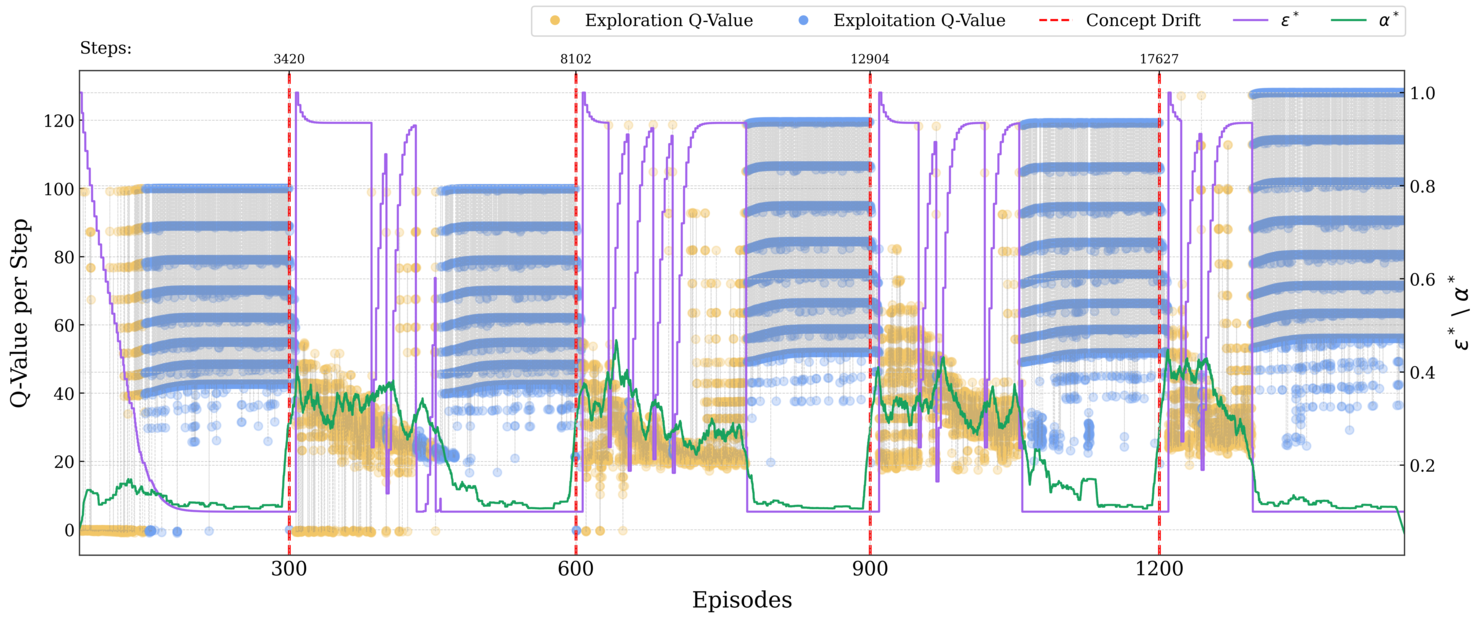}
        \caption{\adaptiverl with adaptive $\varepsilon^*$ and $\alpha^*$.}
        \label{fig:morphin-dynamics}
    \end{subfigure}
    \hfill 
    \begin{subfigure}[b]{0.49\textwidth}
        \centering
        \includegraphics[width=\linewidth]{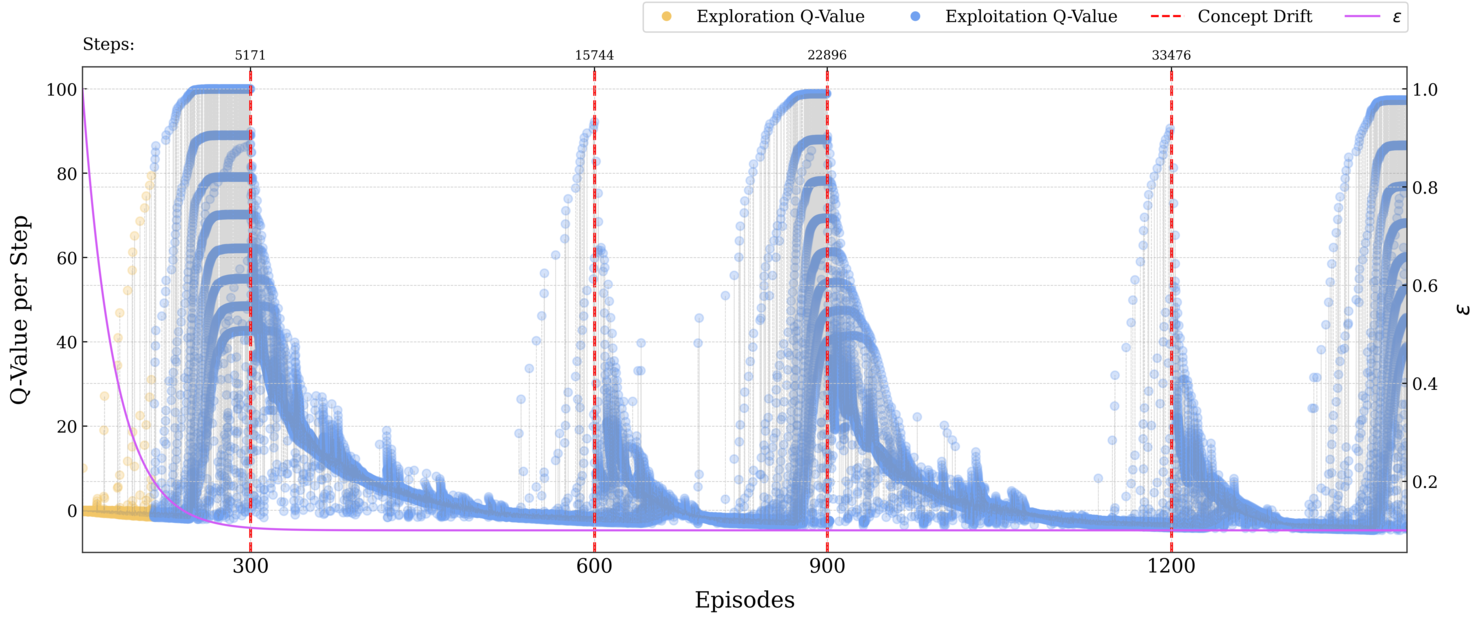}
        \caption{Standard Q-learning with static $\varepsilon$-decay.}
        \label{fig:baseline-dynamics}
\end{subfigure}
    
    \caption{Internal dynamics of \adaptiverl versus a standard Q-learning agent, illustrating their contrasting responses to concept drift (red dashed lines) in the Gridworld scenario, the y-axis for the scatter points represents the Q-value of the chosen state-action pair for that step, while their color indicates whether the action was exploratory (yellow) or exploitative (blue). (a) \adaptiverl's adaptive response: Upon detecting a drift, the PH-test triggers a reset of the exploration rate $\varepsilon^*$ (purple), forcing re-exploration. The resulting high TD-errors cause the dynamic learning rate $\alpha^*$ (green) to increase, accelerating the integration of new knowledge and enabling rapid policy recovery. (b) Standard Q-learning's static behavior: The baseline agent uses a single, exponentially decaying exploration schedule ($\varepsilon$) and a fixed $\alpha$ value. After the initial convergence, the low exploration rate prevents it from adapting to subsequent drifts, causing the agent to remain committed to an obsolete policy and resulting in a sustained performance collapse.}
    \label{fig:dynamics-combined}
\end{figure*}

To adapt, an agent must first recognize environmental changes. \adaptiverl achieves this by monitoring the stream of cumulative episode rewards ($R_{ep}$) with the PH-test~\cite{mignon2017adaptive,networkdynamicrl,changingpointdetection}. A drift is flagged if the cumulative difference between $R_{ep}$ and its running mean exceeds a threshold $H$. The test's sensitivity is controlled by $H$ and a second hyperparameter, $\delta$, which were selected empirically based on the reward scale of each environment. As shown in Algorithm \ref{alg:morphin} (lines 21--28), a detected drift triggers an immediate adaptive response: the exploration decay counter $e$ is reset to zero, forcing the exploration rate $\varepsilon^*$ to its maximum value. This compels the agent to re-explore the environment and learn the new dynamics, as visualized in \fref{fig:morphin-dynamics}.

Once a drift is detected, \adaptiverl employs a two-points strategy. The first part is the mandatory re-exploration described above. The second is a dynamic adjustment of the learning rate $\alpha^*$ to control the speed of knowledge acquisition. This dual mechanism is a cornerstone of \adaptiverl, designed to preserve prior knowledge and prevent catastrophic forgetting. Unlike approaches that retrain from scratch, \adaptiverl never resets the Q-table; existing Q-values serve as an informed starting point for learning the new policy. This reuse of knowledge enables faster adaptation, as evidenced by the shorter re-learning periods showed in \fref{fig:morphin-dynamics}.

The learning rate adaptation is driven by the Temporal Difference (TD) error, which quantifies the discrepancy between the predicted and actual outcomes of an action:
\begin{equation} \label{eq:td_error}
    TD_{error} = r_{t+1} + \gamma \cdot \underset{a}{\max} Q(s_{t+1}, a) - Q(s_t, a_t)
\end{equation}
A concept drift leads to large TD-errors, signaling a mismatch between the agent's knowledge and the new reality. \adaptiverl harnesses this signal to compute a dynamic learning rate $\alpha^*$:
\begin{equation}
    \label{eq:dynamic_learning_rate}
    \alpha^* = \alpha + (\alpha_{\max}-\alpha) \cdot \frac{1}{1 + e^{-(|TD_{error}|-k)}}
\end{equation}
Here, the hyperparameter $k$ controls the sensitivity of $\alpha^*$ to the TD-error and was empirically tuned. High TD-errors yield a large $\alpha^*$, accelerating learning. As the policy stabilizes, TD-errors diminish and $\alpha^*$ decays towards its base value $\alpha$, ensuring convergence.

This framework also handles an expanding action space. When a new action $a_{new}$ becomes available, the Q-table (a matrix of size $|A| \times |S|$) is dynamically augmented by adding a new row for $a_{new}$, initialized to zero. This event, detected as a drift, triggers the same unified response: $\varepsilon^*$ is reset to explore the new action's utility, and $\alpha^*$ adapts to integrate it into the policy.

The effectiveness of \adaptiverl becomes evident when contrasted with a standard Q-learning agent (\fref{fig:baseline-dynamics}). The baseline agent's static exploration schedule prevents adaptation to new goals, causing a sustained drop in performance. In sharp contrast, \adaptiverl (\fref{fig:morphin-dynamics}) explicitly detects changes and coordinates exploration ($\varepsilon^*$) with learning speed ($\alpha^*$) to rapidly converge to a new optimal policy after each drift.

%% file: validation.tex

\section{Validation}
\label{sec:validation}

We evaluate \adaptiverl in two scenarios: a canonical Gridworld benchmark for a clear proof-of-concept, and a more real application over traffic-signal control simulation to validate its adaptability in a scenario representative of real-world self-adaptive systems.

\subsection{Evaluation Scenarios}
Gridworld: In a $9 \times 9$ grid, we run two experiments for 1,000 independent trials each. First, in a 1,500-episode run, the high-reward goal state is swapped between opposite corners every 300 episodes. Second, in a 400-episode run, a new ``jump'' action is introduced at episode 300, requiring the agent to find a more optimal path.

Traffic-Signal Control: We model a two-lane intersection using a custom extended Gym environment~\cite{gymlib}, a well known application for SAS~\cite{HENRICHS2022106940}. The state $s_t = (c_1, c_2)$ is the number of queued vehicles per lane. Actions are signal phases with different service capacities. The reward function penalizes both congestion (queues exceeding a threshold) and inefficiency (allocating green time to empty lanes). Concept drift is induced by changing vehicle arrival rates ($\lambda_1, \lambda_2$) at predefined episodes. Upon drift detection, the action space is expanded with more aggressive signal phases to manage heavier traffic.

\subsection{Experimental Setting}
Experiments were run using Python 3.12 and Gym 0.26.2. We compare \adaptiverl against a Standard Q-learning (Baseline) agent with a fixed learning rate ($\alpha=0.1$) and a single, non-resetting exponential exploration decay. \adaptiverl uses the PH-test to trigger exploration resets ($\varepsilon \to 1$) and a dynamic learning rate $\alpha^*$ modulated by the TD-error. Hyperparameters for both scenarios (e.g., $k=5, \delta=0.5, H=300$) were determined empirically to suit the reward scale of each environment.
\subsection{Evaluation Results}
\begin{figure*}[h]
    \centering
    \begin{subfigure}[b]{0.48\textwidth}
        \centering
        \includegraphics[width=\linewidth]{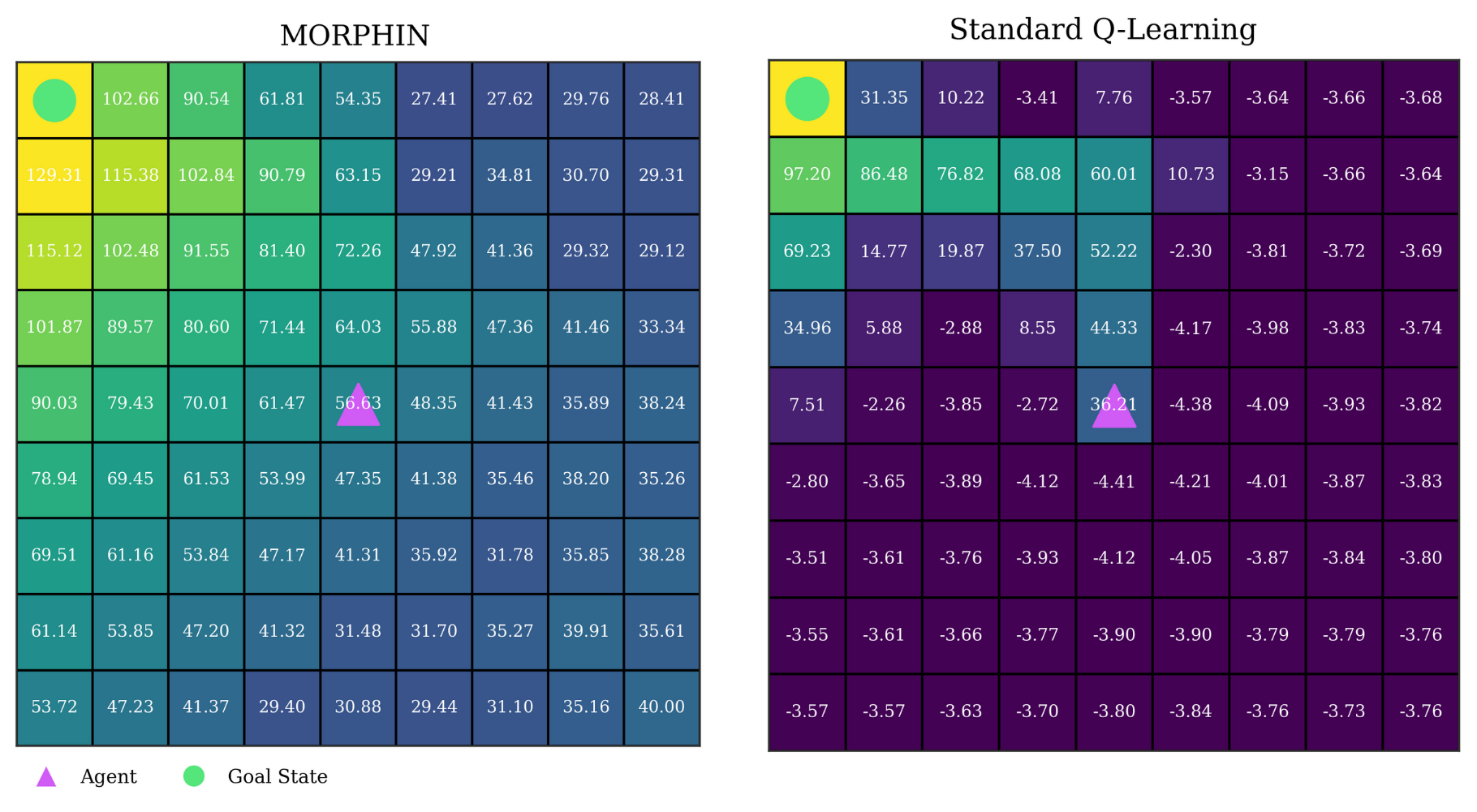}
        \caption{Adaptation to shifting goals (1,500 episodes).}
        \label{fig:q-heatmaps-goals}
    \end{subfigure}
    \hfill 
    \begin{subfigure}[b]{0.48\textwidth}
        \centering
        \includegraphics[width=\linewidth]{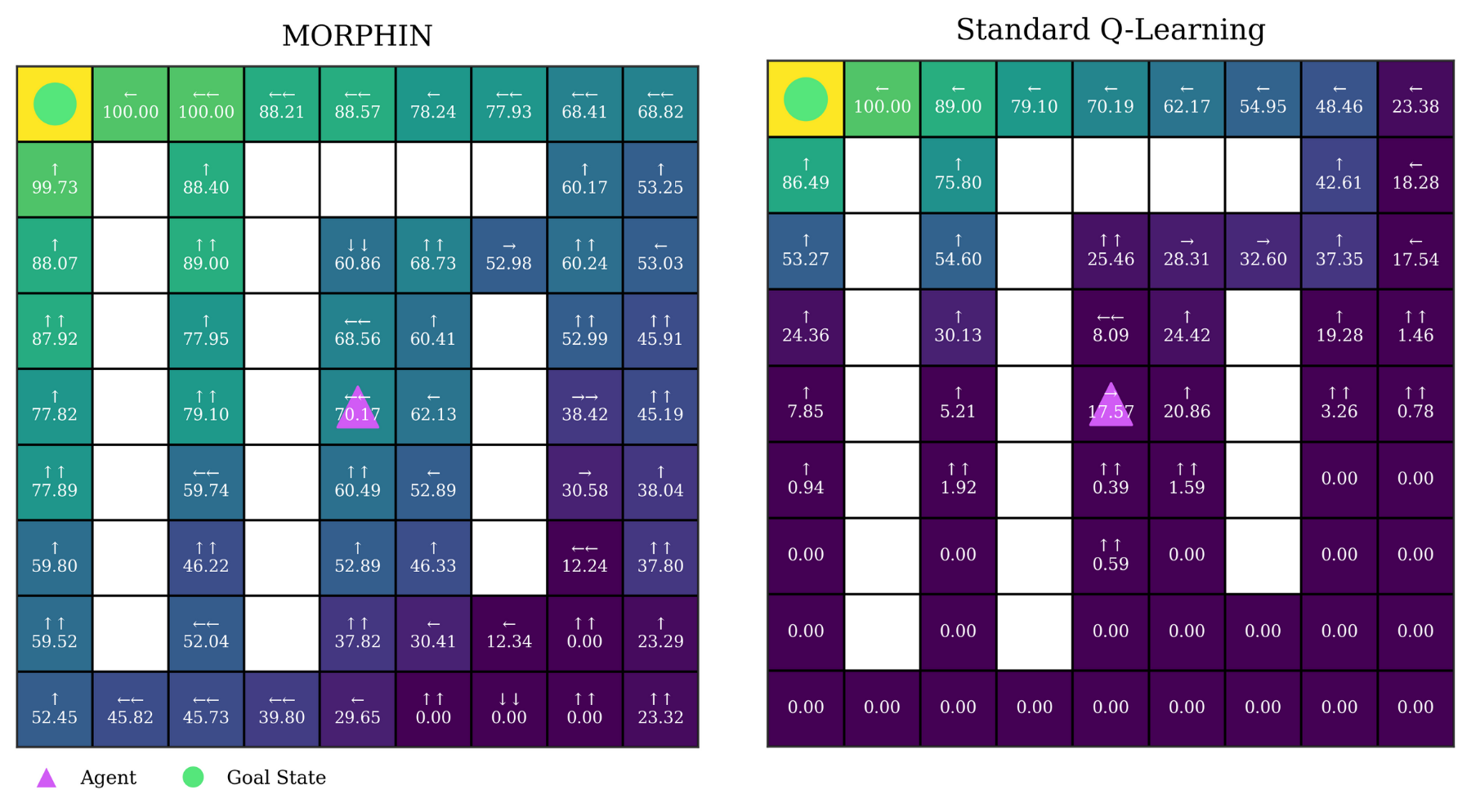}
        \caption{Adaptation to an expanded action space (400 episodes).}
        \label{fig:q-heatmaps-actions}
    \end{subfigure}
    
    \caption{Comparison of Q-value heatmaps demonstrating \adaptiverl's superior adaptation over standard Q-learning in two non-stationary scenarios. (a) In a goal-switching environment, \adaptiverl (left) preserves high Q-values for both initial and subsequent goals, showcasing effective knowledge retention and preventing catastrophic forgetting, unlike the baseline agent (right). (b) When the action space is expanded, \adaptiverl (left) successfully integrates a new ``jump'' action (indicated by double arrows) to discover a more optimal policy, while the baseline (right) fails to adapt, remaining committed to a suboptimal policy.}
    \label{fig:q-heatmaps-combined}
\end{figure*}

Gridworld: \adaptiverl demonstrates superior adaptation and knowledge retention. As illustrated in \fref{fig:q-heatmaps-goals}, after 1,500 episodes, \adaptiverl (left) successfully retains high Q-values for both current and previously learned goals, effectively reducing catastrophic forgetting effects. In contrast, the baseline agent (right) overwrites past knowledge and only remembers the most recent goal. This enhanced adaptability translates to significant efficiency gains, as shown in \fref{tab:gridworld-table}. \adaptiverl improves overall learning efficiency by a factor of 1.7x (measured in total steps) and successfully converges after each induced drift. The baseline agent, however, fails to converge after the first 300-episode interval. Furthermore, \fref{fig:q-heatmaps-actions} shows that when the action space is expanded, \adaptiverl effectively incorporates the new ``jump'' action to discover a more optimal policy, whereas the baseline agent struggles to update its established policy and remains suboptimal.

\begin{figure*}[h]
    \centering
    \begin{minipage}{0.7\textwidth}
        \includegraphics[width=\textwidth]{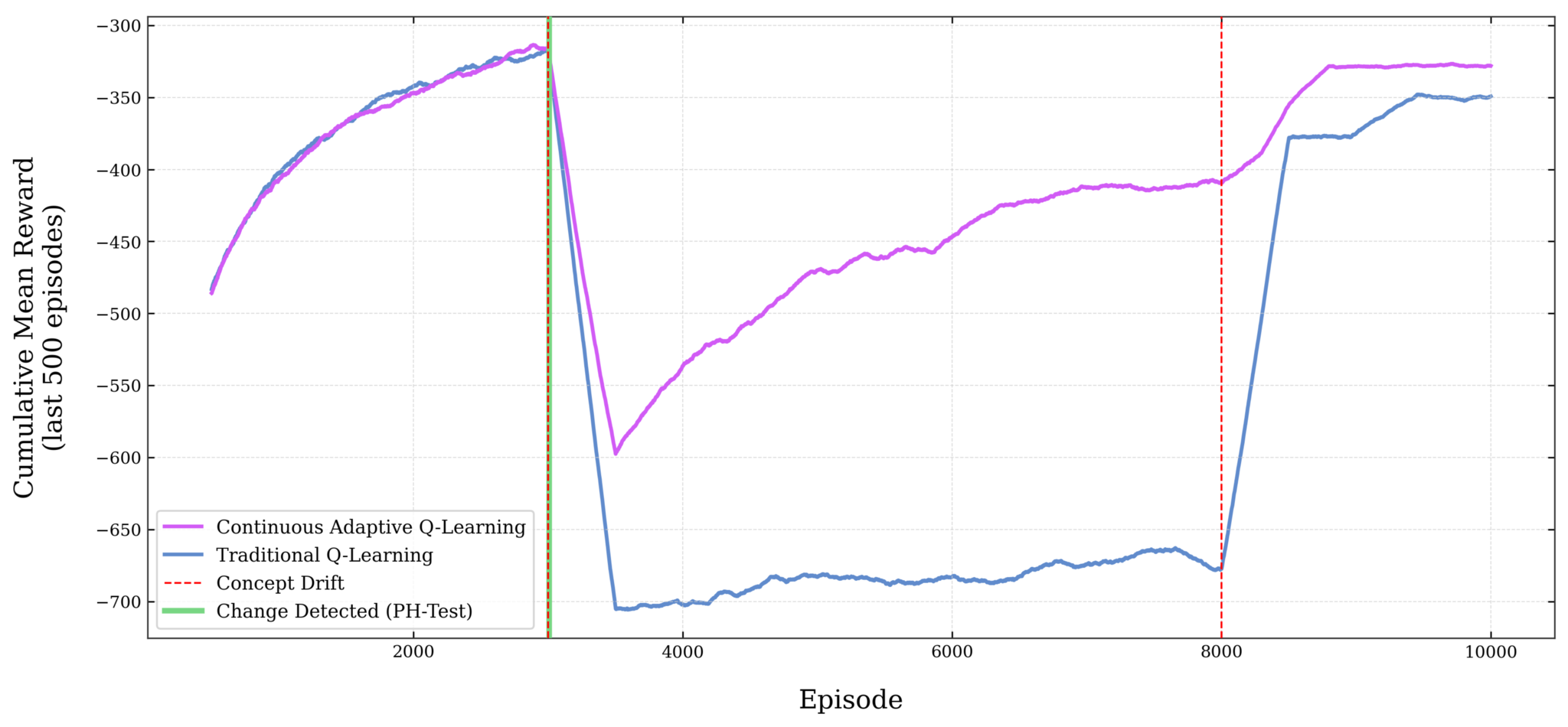}
    \end{minipage}
    \hfill
    \begin{minipage}{0.25\textwidth}
        \caption{Learning performance in the traffic scenario. After the first drift (episode 3,000), \adaptiverl detects the change (green line) and rapidly recovers. Traditional Q-learning suffers prolonged degradation. The second drift (episode 8,000) is not detected by the PH-tests because the new reward distribution is a subset of previously seen values and does not exceed the sensitivity threshold.}
        \label{fig:traffic-learning-curve}
    \end{minipage}
\end{figure*}

Traffic-Signal Control: In this scenario, \adaptiverl again shows robust adaptation (\fref{fig:traffic-learning-curve}). When congestion increases at episode 3,000, the PH-test detects the drift, triggering adaptation and enabling a rapid performance recovery. The baseline agent suffers a prolonged degradation. However, the results also highlight a limitation: a second drift at episode 8,000 (lowered traffic) is not detected because the new reward distribution is a subset of previously seen values and does not exceed the PH-test's sensitivity threshold. While both agents performance improves in the easier environment, this failure underscores the challenge of parameterizing drift detectors. Despite this, the case study confirms \adaptiverl's ability to react to detected non-stationarity and seamlessly incorporate new actions to manage changing conditions.

\begin{table}[h]
\centering
\caption{Average convergence time (in episodes after drift) and total steps over 1,500 episodes in Gridworld (1,000 runs). Dashes (--) indicate failure to converge within the 300-episode interval. An independent t-test confirms the difference in total steps is statistically significant ($p < 0.05$).}
\label{tab:gridworld-table}
\resizebox{\columnwidth}{!}{
\begin{tabular}{l | c | c | c | c }
\toprule
\textbf{Agent} & \textbf{1st Drift} & \textbf{2nd Drift} & \textbf{3rd Drift} & \textbf{Total Steps} \\
\midrule
Q-Learning  & 256.40 $\pm$ 4.35\% & -- & -- & 40,683.74 $\pm$ 1.33\% \\
\textbf{\adaptiverl} & 135.81 $\pm$ 9.31\% & 175.17 $\pm$ 5.80\% & 167.81 $\pm$ 3.13\% & 23,292.07 $\pm$ 6.33\% \\
\bottomrule
\end{tabular}
}
\end{table}

%% file: related.tex

\section{Related Work}
\label{sec:related}

Learning in non-stationary environments is a significant challenge in \ac{RL}, revolving around the exploration-exploitation trade-off~\cite{sutton18}. Our work builds on established research lines addressing this issue, comprehensively surveyed by \citet{Padakandla_2021}.

\subsection{Adapting to Environmental Changes}

A common strategy in non-stationary environments is adapting learning hyperparameters. Several works adjust the exploration rate, $\varepsilon$, often using a change-point detector to trigger the adaptation. Both~\citet{mignon2017adaptive} and~\citet{changingpointdetection} used the PH-test to detect drifts and adapt exploration, a combination our work adopts. Other model-free approaches include Repeated Update Q-learning (RUQL), which addresses policy-bias by repeating updates for less-chosen actions~\cite{nonstationaryqlearning}, contrasting our method's adaptation of the learning rate $\alpha^*$ based on TD-error magnitude. Proactive methods like RestartQ-UCB periodically reset memory on a fixed schedule~\cite{pmlr-v139-mao21b}, whereas \adaptiverl is reactive, using an online PH-test to trigger adaptation only when performance changes significantly. Context-based methods like Context Q-learning~\cite{Padakandla_2020} maintain separate Q-tables for each environmental context, isolating knowledge at the cost of increased memory. \adaptiverl follows a more lightweight philosophy, maintaining a single Q-table and adapting in-place by dynamically scaling the learning rate to integrate new experiences.

\subsection{Continual Learning and Self-Adaptive Systems}
The aforementioned methods align with the principles of \ac{CRL}~\cite{khetarpal2022continualreinforcementlearningreview}, where the goal is for agents to ``never stop learning''~\cite{abel2023definitioncontinualreinforcementlearning}. Recent studies show that many CL methods struggle when a state-action pair yields different rewards after an environmental shift~\cite{Bagus2022}. \adaptiverl addresses this directly: by not resetting its Q-table and forcing re-exploration, an outdated Q-value generates a large TD-error when encountering a new reward, which in turn drives a high learning rate to rapidly overwrite the obsolete knowledge. \ac{RL} is also a potent tool for building self-adaptive systems~\cite{HENRICHS2022106940}, with applications in IoT security~\cite{iotdynamicrl} and network monitoring~\cite{networkdynamicrl}. Our traffic-control scenario is a canonical example of this paradigm.

\subsection{Positioning of Our Contribution}
The main contribution of this paper is not a single algorithmic component, but rather the synthesis and integration of several established techniques into a unified, model-free framework for tabular Q-learning. \adaptiverl combines proactive environment monitoring via the PH-test with reactive adaptation of both the exploration rate ($\varepsilon^*$) and the learning rate ($\alpha^*$). This coordinated strategy is specifically designed to enable an agent to adapt to two distinct types of non-stationarity simultaneously: changes in the reward function (shifting goals) and on-the-fly expansions of the action space. By preserving and building upon a single Q-table, our approach provides a lightweight solution aimed at preventing catastrophic forgetting and ensuring continual adaptation in dynamic environments, without the need for full retraining or maintaining multiple explicit context models.

%% file: conclusion.tex

\section{Conclusion and Future Work}
\label{sec:conclusion}

This paper introduced \adaptiverl, a self-adaptive framework for tabular Q-learning agents operating in non-stationary environments. We specifically addressed the challenge of agents adapting to simultaneous changes in their goals (reward functions) and their available capabilities (action-space expansions). Our approach integrates concept drift detection using the PH-test with dynamic adjustments to the exploration ($\varepsilon$) and learning ($\alpha$) rates. Experimental results in both a Gridworld benchmark and a traffic control simulation demonstrate that this coordinated strategy enables agents to adapt more effectively than a standard Q-learning baseline, achieving a performance increase of up to $1.7\times$ in learning efficiency while successfully reducing catastrophic forgetting effects.

Through this synthesis of established techniques into a unified framework, \adaptiverl demonstrates how a lightweight, model-free agent can achieve robust continual learning. By preserving and adapting a single Q-table, it effectively enables knowledge reuse when faced with contradictory environmental changes, a key challenge highlighted by~\citet{Bagus2022}, without requiring full retraining or multiple context models. Our work thus presents a practical and resource-efficient method for developing self-adaptive systems capable of real-time learning in dynamic environments.

This work represents an initial validation, and several avenues for future work are evident. The limitations observed in our experiments, such as the failure of the PH-test to detect certain drifts and the need for empirical tuning of hyperparameters, point to clear directions for improvement. Future work should therefore focus on:
\begin{enumerate}
    \item Generalization and Scalability: Extending the core principles of \adaptiverl from tabular methods to deep \ac{RL} architectures to handle high-dimensional state spaces. This would also involve investigating the use of memory-based plasticity to enhance knowledge transfer across tasks.
    \item Robustness and Empirical Analysis: Conducting a rigorous sensitivity analysis of the introduced hyperparameters, particularly the TD-error sensitivity ($k$) and $H$ threshold for the PH-test, to better understand their impact on performance. Furthermore, we plan to explore more robust drift detection methods to overcome the limitations of the PH-test observed in our traffic scenario, where drifts that result in subsets of known reward distributions can be missed. We also intend to extend the evaluation to include scenarios with action removal and other types of reward function changes, such as those examined by \citet{mignon2017adaptive}.
    \item Advanced Application Domains: Applying the framework to more complex distributed multi-agent systems (\eg bigger street network configuration). Finally, we plan to deploy \adaptiverl on resource-constrained edge devices for real-world applications in \ac{IOT} and robotics.
\end{enumerate}